\documentclass[runningheads]{llncs}
\usepackage{graphicx}
\usepackage{amsmath,amssymb} 
\newcommand\norm[1]{\left\lVert#1\right\rVert}
\usepackage{color}

\begin{document}
%
\title{Large-Scale Video Classification with Feature Space Augmentation coupled with Learned Label Relations and Ensembling}

\titlerunning{Large-Scale Video Classification with Feature Space Augmentation}
%
\author{Choongyeun Cho\inst{1} \and
Benjamin Antin\inst{1,2} \and
Sanchit Arora \and
Shwan Ashrafi\inst{1} \and
Peilin Duan\inst{1} \and
Dang The Huynh\inst{1} \and
Lee James\inst{1} \and
Hang Tuan Nguyen\inst{1} \and
Mojtaba Solgi\inst{1} \and
Cuong Van Than\inst{1}}
%
\authorrunning{C. Cho et al.}
%

\institute{Axon, 1100 Olive Way, Seattle WA 98004, USA \and
Stanford University, 450 Serra Mall, Stanford CA 94305 USA \and
\email{cycho@axon.com, \\
bantin@stanford.edu, \\
sanchitarora13@gmail.com, \\
\{sashrafi,pduan,dhuynh,ljames,hnguyen,msolgi,cthan\}@axon.com}}
\maketitle              

\begin{abstract}
This paper presents the Axon AI's solution to the 2nd YouTube-8M Video Understanding Challenge, achieving the final global average precision (GAP) of 88.733\% on the private test set (ranked 3rd among 394 teams, not considering the model size constraint), and 87.287\% using a model that meets size requirement.
Two sets of 7 individual models belonging to 3 different families were trained separately.
Then, the inference results on a training data were aggregated from these multiple models and fed to train a compact model that meets the model size requirement.
In order to further improve performance we explored and employed data over/sub-sampling in feature space, an additional regularization term during training exploiting label relationship, and learned weights for ensembling different individual models.
\end{abstract}
%
%
%
\section{Introduction}
Video classification and understanding is an emerging and active area of research as a video domain may be the fastest growing data source in the last decade.
Yet the video classification problem is still largely unsolved and far behind human capability.
One of the reasons for this has been a lack of realistic, large-scale video dataset.
YouTube-8M is such a large-scale video dataset with high-quality machine-annotated labels.
It provides pre-extracted audio and visual features computed from millions of YouTube videos for faster data access and training.
The goal of this YouTube-8M Video Understanding Challenge was to develop a machine learning model that accurately predicts labels associated with each unseen test video.

\section{Challenge strategy} \label{sec:strategy}
Our overall plan for the Challenge was as follows:
(1) designing or identifying a set of efficient and strong submodels;
(2) ensembling predictions from all the submodels;
(3) distilling into a model that satisfies the model size constraint (i.e. less than 1GB when uncompressed); and
(4) exploring and implementing various improvement ideas during individual model training or knowledge-distillation training, namely (a) data augmentation, (b) exploiting label relationship, and (c) trying different ensembling methods.

Early in the competition stage, we started to identify powerful and efficient baseline models regardless of their model sizes.
We explored approaches and models from top-performing participants in the last year's Kaggle Challenge~\cite{kaggle2017}, and the model architectures from the 1st place winner \cite{miech2017} turned out to be excellent references as they present high performance (in terms of GAP) and quick training and inference time.
We also observed train, validate and supposedly test datasets are well randomized and balanced so that GAP on the validate set is representative of GAP on the test set, and even very small subset of labeled data can reliably serve as a new ``validate'' data.
In order to increase data samples for training, we used all training data (i.e. \texttt{train????.tfrecord} files in a wildcard notation) and about 90\% of the validate data (i.e. \texttt{validate???[0-4,6-9].tfrecord}) for all the training of single models.
Only one tenth of the validate data (i.e. \texttt{validate???5.tfrecord}) was set aside for training monitoring, model and hyper-parameter selection, and ensemble weight learning.

The predictions from multiple models were aggregated and ensembled in order to enhance the GAP performance.
In a nutshell, all the information about a single model is represented as predictions on a training dataset.
In this way, many single models can be effectively combined without having to run inference of multiple models concurrently.
For ensembling schemes, we implemented (1) simple averaging with equal weights, (2) per-model linearly weighted average, and (3) per-model and per-class linearly weighted average.
Among these ensembling schemes, the per-model weights provided the best performance improvement.

In order to meet the model size requirement for the Challenge, knowledge distillation was performed based on the implementation from the original paper~\cite{hinton2014} using only the soft targets from a ``teacher'' model (an ensemble of multiple submodels), and not the ground truth (hard targets).

We experimented other improvement ideas which will be described in the subsequent subsections.

\subsection{Single baseline models}
We took advantage of three model families depending on the pooling strategy to aggregate frame-level representations into a global, video-level representation, namely: learnable pooling (LP), bag of words (BoW), and recurrent neural network (RNN) models from the last year's winning method~\cite{miech2017}.

LP method encodes the frame-level features using Fisher vectors (FV) or some of its simplified variants including vector of locally aggregated descriptors (VLAD), residual-less VLAD (RVLAD).
For all these variations of LP method, the cluster centers and soft assignments are learned in an end-to-end fashion.

``Gated'' version of each model utilizes context gating, the learned element-wise multiplications (gates) at the last layer of a model:
\begin{equation}
y = \sigma (W \cdot x + b) \circ x
\end{equation}
where $x$ and $y$ are input and output feature vectors respectively, $\sigma$ is an element-wise sigmoid function, and $\circ$ means element-wise multiplication.

\begin{table}[h!]
  \begin{center}
    \caption{A set of 7 single baseline models before ensembling}
    \label{tab:single_models}

    \begin{tabular}{ c | c }
      Model family & Brief description of individual models  \\
    \hline
    \hline
            LP & Gated NetVLAD with 256 clusters \\
    \hline
            LP & Gated NetFV with 128 clusters \\
    \hline
            BoW & Gated soft-DBoW with 4096 clusters \\
    \hline
            BoW & Soft-DBoW with 8000 clusters \\
    \hline
            LP &  Gated NetRVLAD with 256 clusters \\
    \hline
            RNN & Gated recurrent unit (GRU) with 2 layers and 1024 cells per layer \\
    \hline
            RNN & LSTM with 2 layers and 1024 cells per layer \\
    \end{tabular}
  \end{center}
\end{table}

Table \ref{tab:single_models} lists a brief description of a set of 7 submodels later used in an ensembled model, roughly in the order of decreasing GAP accuracy (hence, Gated NetVLAD being the most powerful and LSTM being the least).
As in the original approach of Willow team, we utilized all 7 models as they represent diverse model architecture families and are known to perform very competitively.
In the end, we used two sets of these 7 models, totaling 14 single models.

All the single models are trained with the cross-entropy loss as it is known to work well for the performance metric of choice, GAP.
The source code for these models was publicly available ~\cite{miech_github}.

\section{Experiments}
Table~\ref{tab:gap} shows progressive improvements of classification accuracy in terms of GAP (all evaluated on a test set unless stated otherwise)

The strongest single model was gated NetVLAD achieving 85.75\%.
This model was combined (simple averaging) with video-level 16-expert MoE model trained with augmented dataset to achieve 0.23\% gain (see subsection~\ref{sub:augmentation}).

From this trained model, another 40,000 iterations were trained with additional regularized term that exploits label relationship (see subsection~\ref{sub:label_relationship}), to reach 87.88\%.

Simple averaging of one or two sets of 7 models offered 88.27\% and 88.62\% respectively (see subsection~\ref{sub:ensemble}).
Learned weights per model instead of equal weights for all models, provided additional 0.11\% improvement.

After teacher-student knowledge distillation (see subsection~\ref{sub:distillation}) we achieved 87.32\%.

\begin{table}[h!]
  \begin{center}
    \caption{GAP performance per experiment}
    \label{tab:gap}
    \begin{tabular}{l|c} 
            \textbf{Experiment} & \textbf{Test GAP (\%)} \\
      \hline
      \hline
            Single baseline model (gated NetVLAD) & 85.75 (Val GAP) \\
      \hline
            Single gated NetVLAD model + video-level MoE model \\
            trained with augmented dataset in feature space & 85.98 (Val GAP) \\
      \hline
            Single gated NetVLAD model + regularized DNN \\
            exploiting label relationship & 87.88 (Val GAP) \\
      \hline
      A simple average ensembling of \\ all of the 7 models & 88.27 \\
      \hline
      A simple average ensembling of \\ two sets of all of the 7 models \\ (14 models in total) & 88.62  \\
      \hline
            Ensembled using learned weights &  \textbf{88.73} \\
      \hline
            Distilled model & \textbf{87.32} \\
    \end{tabular}
  \end{center}
\end{table}

\subsection{Data over- and sub-sampling} \label{sub:augmentation}
Train dataset augmentation is an effective way to increase data samples for training, hence potentially improving generalizability and performance of a classifier, without having to explicitly annotate additional data.
A common practice is to apply a small perturbation in the original data domain (cropping, mirroring, color jittering in the case of image domain, for example).

\begin{figure}
  \includegraphics[width=\linewidth]{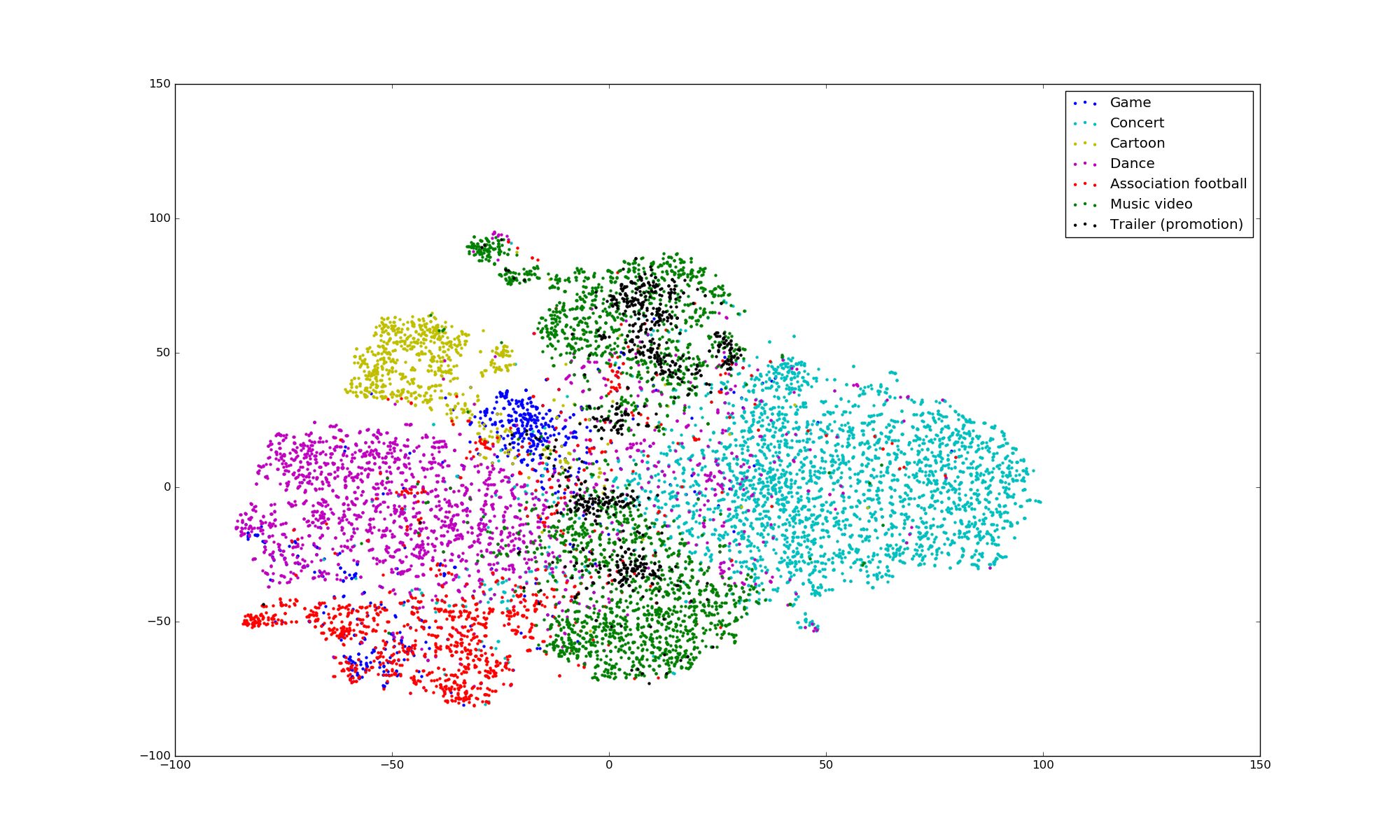}
  \caption{A TSNE plot of visual features for a few selected classes (best viewed in color)}
  \label{fig:tsne}
\end{figure}

Figure~\ref{fig:tsne} shows TSNE visualization of visual features for several selected classes.
Note that data examples belonging to only single label have been plotted in this figure for the sake of easier visualization.
It is observed that examples (videos) associated with a same semantic concept (label) form a cluster while videos belonging to different concepts are separated to some degree.

\begin{figure}
  \includegraphics[width=\linewidth]{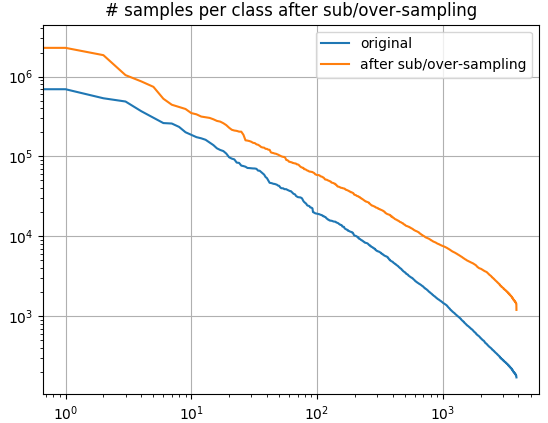}
  \caption{Label counts before and after data augmentation in feature space}
  \label{fig:numsamples}
\end{figure}

The label frequencies (counts) are plotted in log-log scale Figure \ref{fig:numsamples}.
It is clear that the plot follows a Zipf distribution: relatively few classes dominate the number of examples, and the tail distribution is very ``fat.''
In order to cope with this distribution both over- and sub-sampling were employed.
With data augmentation (over-sampling) we hoped to fill in the gap inside a cluster of same label especially for those classes with fewer examples.
In addition to over-sampling, sub-sampling (random sampling) for classes with more than enough examples will make training set more balanced and expedite the training time.

In YouTube-8M dataset we are provided with pre-extracted features.
As for the video-level visual features, PCA, whitening and quantization have been performed on a Inception-model deep feature (DNN output) per frame, and all the frame-level features are averaged.

Data augmentation was performed inspired by \cite{devries2017} on the video-level visual features due to memory and computation limitations.
The simplest transform is to simply add small noise to the feature vector.
\begin{equation}
x'_i = x_i + \gamma Z, Z \sim \mathcal{N}(0, \sigma^2)
\end{equation}

For each sample in the dataset, we find its $K$ nearest neighbors in feature space (in L2 sense) which share its class label.
For each pair of neighboring feature vectors, a new feature vector can then be generated using interpolation:
\begin{equation}
x'_i = x_i + \lambda_i (x_j - x_i)
\end{equation}
where $x'_i$ is the synthesized feature vector, $x_j$ is a neighboring feature vector to $x_i$ and $\lambda_i$ is a parameter in the range of 0 to 1 which dictates the degree of interpolation.

In a similar fashion, extrapolation can also be applied to the feature vectors:
\begin{equation}
x'_i = x_i + \lambda_e (x_i - x_j)
\end{equation}
We used 0.5 for both $\lambda$'s and $\sigma$ value of 0.03 for the additive Gaussian noise.

The label frequencies (before and after over- and sub-sampling) are plotted in log-log scale Figure \ref{fig:numsamples}, both of which exhibit a Zipf-like distribution.

There were some implementation considerations because of limited memory and storage resources: Instead of finding ``global'' nearest neighbors over all examples for a given label (scanning all tfrecord files), we limited a number of tfrecord files to search within, to 256 files at a time, effectively processing data augmentation in 256 tfrecord chunks.
Also, since YouTube8M dataset is inherently multi-labeled, the decision of whether and how much over-sampling and sub-sampling will be done given an example is based on a single label which is least frequently occurring over all labels.
That is, if a video is associated with a single label which is in a sub-sampling regime (i.e. having more than $10^4$ examples which is likely to be more than enough), this video will be subject to random \textbf{sub-sampling}.
If a video is associated with a single label which is in a over-sampling regime (having less than $10^4$ examples), this video will be \textbf{over-sampled} using aforementioned augmentation schemes.
Number of nearest neighbors are selected heuristically based on label frequencies (more nearest neighbors for the labels with fewer examples).
If a video is associated with multiple labels, then the label with the least examples will dictate the over- and sub-sampling decision.

Table~\ref{tab:aug_num_samples} shows number of training examples before and after data augmentation.
Data augmentation more than quadrupled the number of samples.

\begin{table}[h!]
  \begin{center}
    \caption{GAP performance per experiment}
    \label{tab:aug_num_samples}

    \begin{tabular}{ l | c }
       & Number of training examples  \\
    \hline
    \hline
      Before data augmentation & 5,001,275  \\
    \hline
      After data augmentation & 23,590,464 \\
    \end{tabular}
  \end{center}
\end{table}

\subsection{Label relationship} \label{sub:label_relationship}
Utilizing label relationship in multi-label classification setting is actively investigated.
Many of the approaches involve modification of the existing model architecture and explicitly calculating and incorporating co-occurrence matrix \cite{rabinovich2007}\cite{bengio2013}.
Some have explored strict hierarchical relationships among different classes (mutual exclusion and subsumption, for example \cite{deng2014}), but this assumption is not suitable in the case of the YouTube-8M dataset as labels are machine-generated, hence inherently noisy.

Among different approaches to address label relationship, an additional regularized term that takes advantage of class relationship \cite{jiang2018} was implemented.
This method was especially preferred for this Challenge because any model can be first trained in a normal setting without this extra regularization; then, after training matures, the extra regularization can be applied in a fine-tuning setting since the calculation of this regularization term is computationally intensive.
\begin{equation}
        \min_{\boldsymbol{W}, \Omega} \sum_{i=1}^{N} l(f(x_i), y_i) + \frac{\lambda_1}{2} \sum_{l=1}^{L-1} \norm{ \boldsymbol{W}_l }_F^2 + \lambda_2 \cdot \textrm{tr}( \boldsymbol{W}_{L-1} \Omega^{-1} \boldsymbol{W}_{L-1}^T) \\
\end{equation}
\[
        \textrm{s.t. } \Omega \succeq 0
\]
where $x_i$ and $y_i$ are an $i$-th input example and target, $\lambda$'s are regularization coefficients, and $\boldsymbol{W}_l$ is a $l$-th layer model weights (hence, $\boldsymbol{W}_{L-1}$ being the last layer's weights), and $\Omega \in \mathbb{R}^{C \times C}$ encodes label relationship.

The first part in the above cost function measures the empirical loss on the training data, which summarizes the discrepancy between the outputs of the network and the ground-truth labels.
The second part is a regularization term to mitigate overfitting.

The last part imposes a trace norm regularization term over the coefficients of the output layer $\boldsymbol{W}_{L-1}$ with the class relationships augmented as a matrix variable $\Omega$.
The positive semidefinite constraint, $\Omega \succeq 0$ indicates that the class relationship matrix is viewed as the similarity measure of the semantic classes.
The original paper \cite{jiang2018} suggests using alternating optimization algorithm to solve for both $\boldsymbol{W}$ and $\Omega$. After updating $\boldsymbol{W}$ using backpropagation, $\Omega$ can be updated as:
\begin{equation}
        \Omega = \frac
        { ( \boldsymbol{W}_{L-1}^T \boldsymbol{W}_{L-1} )^\frac{1}{2}  }
        { \textrm{tr} (( \boldsymbol{W}_{L-1}^T \boldsymbol{W}_{L-1} )^\frac{1}{2} )  }.
\end{equation}

\subsection{Ensembling} \label{sub:ensemble}
For training of individual baseline models, we stopped training when a model starts to overfit slightly by monitoring validation GAP in the spirit of the finding from \cite{bober2017} and our own observations.
For ensembles, we implemented (1) simple averaging (equal weights for all models), (2) per-model linearly-weighted average, (3) per-model and per-class linearly-weighted average.
Among these ensembling methods, per-model weights provided the best performance improvement.

To learn the per-model weights for ensembling, a dataset was made that comprised of each model's inference results on our validation set (one tenth of the original validation set as described in Section~\ref{sec:strategy}).
Stochastic gradient descent (SGD) with the Adam optimizer was then used to minimize the mean-square-error (MSE) loss between a linear combination of the models' inferences and the ground truth.
A custom initializer was used to make the model converge, typically in less than 10 epochs, on useful weights.
The initializer used a normal distribution with a mean of $1/(\# \textrm{model})$ and standard deviation of 0.05 to approximate weights for MSE.
Learned per-model weights (Table~\ref{tab:learned_weights}) look reasonable as (1) Gated NetVLAD was the strongest model in terms of GAP performance, and as (2) weights are roughly in a decreasing order from most powerful (Gated NetVLAD) to least powerful model (LSTM).

\begin{table}[h!]
  \begin{center}
    \caption{Learned weights for 7 baseline models}
    \label{tab:learned_weights}

    \begin{tabular}{ l | c }
     Model & Weight \\
    \hline
    \hline
            Gated NetVLAD & 0.2367 \\
    \hline
            Gated NetFV & 0.1508 \\
    \hline
            Gated soft-DBoW & 0.1590 \\
    \hline
            Soft-DBoW & 0.1000 \\
    \hline
            Gated NetRVLAD & 0.1968\\
    \hline
            GRU & 0.1306 \\
    \hline
            LSTM & 0.0621
    \end{tabular}
  \end{center}
\end{table}

\subsection{Knowledge distillation} \label{sub:distillation}
We used the predictions of the ensemble model $\tilde{p}$ as soft targets along with the ground truth targets $q$ for training a student model.
The loss function can be written as the weighted sum of two cross-entropy losses ($CE(\cdot,\cdot)$) as
\begin{equation}
L = \lambda \cdot CE(p, \tilde{p}) + (1-\lambda) \cdot CE(p,  q),
\end{equation}
where $p$ is the predictions of the student model.
We trained the student model using different values of $\lambda$’s, and the best GAP result was achieved with $\lambda = 1$, i.e. pure distillation without using the ground truth targets.

The choice of the student model was based on two factors (1) the 1GB constraint on the size of the final model and (2) the best GAP number one could expect from the candidate single models.
As such we chose to use the gated NetVLAD model for it had the best performance amongst the single models as reported by \cite{miech2017}.
However, the NetVLAD with 1024 hidden weights in the last fully connected layer results in a model size greater that the 1GB limit.
Therefore, the number of hidden weights was reduced to 800 which yielded in a model size slightly less that the limit.

\subsection{Training details}
We kept training details unchanged from the original implementation of these models \cite{miech2017}.
All models are trained using the Adam optimizer.
The learning rate is initialized to 0.0002 and is exponentially decreased with the factor of 0.8 for every 4M examples.
For all the clustering-based pooling methods (NetVLAD, NetRVLAD, NETFV, and Soft-DBoW), 300 frames were randomly sampled with replacement.


\section{Conclusions}
We approached this YouTube-8M Video Understanding Challenge with a clear and methodical planning and strategy, and achieved 88.733\% final GAP (ranked the 3rd place, not considering the model size constraint), and 87.287\% using a valid model that meets size requirement.
In addition to identifying and employing strong baseline classifiers, we implemented data augmentation in feature space, an extra regularization term that exploits label relationship, and learned weights for the ensembling.

\section{Acknowledgement}
The authors would like to thank Youtube-8M Challenge organizers for hosting this exciting competition and for providing the excellent starter code, and the Axon team to support this project.

\bibliographystyle{splncs04}
\bibliography{W46P18}
\end{document}